\pdfoutput=1
\documentclass[10pt,twocolumn]{article}

\usepackage[letterpaper,margin=0.75in]{geometry}
\usepackage{microtype}
\usepackage{amsmath,amsfonts}
\usepackage{amssymb}
\usepackage{mathtools}
\usepackage{array}
\usepackage{booktabs}
\usepackage[table]{xcolor}
\usepackage{colortbl}
\usepackage{graphicx}
\usepackage{textcomp}
\usepackage{url}
\usepackage{cite}
\usepackage{pifont}
\usepackage{caption}

\definecolor{ourscolor}{HTML}{E8F0E8}

\usepackage[colorlinks=true,linkcolor=blue,citecolor=blue,urlcolor=blue]{hyperref}
\hypersetup{
    pdftitle={CARE: Competence-Aware Reward Shaping for Adaptive Reasoning Length in Video-MLLMs},
    pdfauthor={Chengwen Liu, Hao Peng, Jisheng Dang, Hong Peng, Bin Hu, Tat-Seng Chua}
}

\title{CARE: Competence-Aware Reward Shaping for\\Adaptive Reasoning Length in Video-MLLMs}

\author{
Chengwen Liu$^{1}$ \quad
Hao Peng$^{1}$ \quad
Jisheng Dang$^{1,\dagger}$ \quad
Hong Peng$^{1,\dagger}$ \quad
Bin Hu$^{2,\dagger}$ \quad
Tat-Seng Chua$^{3}$\\[0.5em]
\small $^1$School of Information Science and Engineering, Lanzhou University, Lanzhou, China\\
\small $^2$School of Medical Technology, Beijing Institute of Technology, Beijing, China\\
\small $^3$School of Computing, National University of Singapore, Singapore\\[0.5em]
\small $^\dagger$Corresponding authors: Jisheng Dang, Hong Peng, and Bin Hu.
}
\date{}

\begin{document}

\maketitle

\begin{abstract}
In multimodal video reasoning, reinforcement learning-based methods typically rely on simplistic and inflexible reasoning-length control strategies that fail to adapt to the model's evolving competence. This mismatch may suppress necessary exploration at early stages, while encouraging redundant reasoning and inefficient decoding once the model becomes more competent. In this paper, we propose \textbf{CARE}, a \textbf{c}ompetence-\textbf{a}ware \textbf{r}eward \textbf{s}haping framework for adaptive reasoning length optimization in multimodal reasoning. Specifically, CARE maintains a smoothed competence estimate via an exponential moving average of pass rates, and uses it to route training into progressive stages that shift the reward preference from exploration-oriented long-form reasoning to efficiency-oriented concise reasoning. To avoid conflating verbosity with intrinsic task complexity, CARE further normalizes reasoning effort with batch-level statistics, and introduces a posterior amplifier to strengthen reward signals for unexpectedly strong performance on historically difficult samples. The proposed mechanism is seamlessly integrated into the GRPO training pipeline and incurs no additional inference-time overhead. Extensive experiments on multiple video reasoning and general video understanding benchmarks demonstrate that CARE consistently improves reasoning accuracy, stabilizes reinforcement learning, and significantly enhances token efficiency. Moreover, CARE exhibits a characteristic inverted-U trajectory of reasoning length during training, and yields shorter yet more informative reasoning traces at convergence, indicating effective adaptive allocation of reasoning budget. We provide the source code for our proposed CARE framework and experiments at \url{https://github.com/1Pansy/Video-CARE}.
\end{abstract}


\begin{figure*}[t]
\centering
\includegraphics[width=\textwidth]{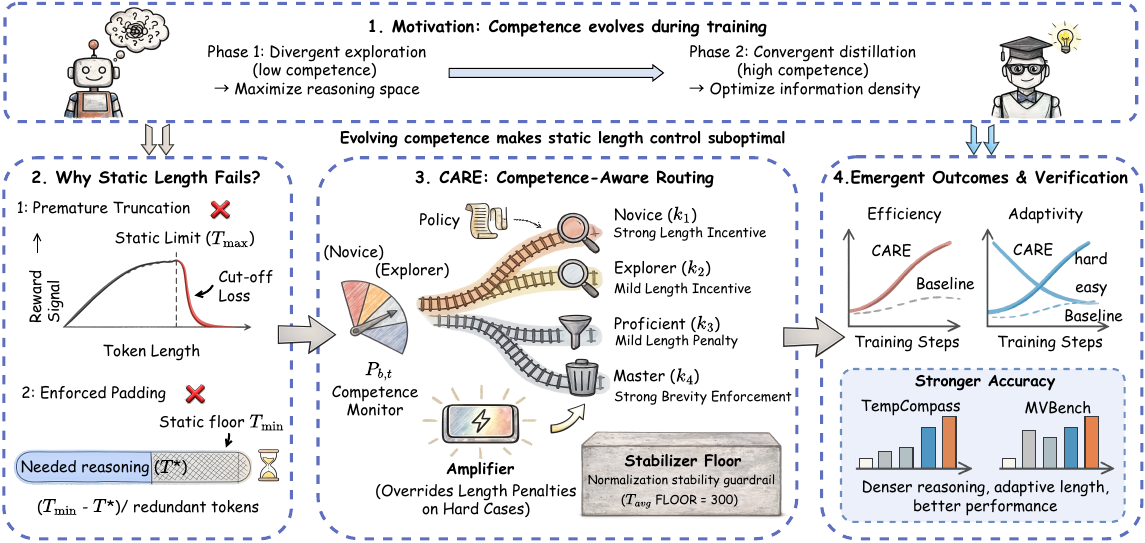}
\caption{Motivation of CARE. As model competence evolves during reinforcement learning, a fixed preference over reasoning length becomes increasingly mismatched to the policy's needs: it can truncate useful exploration early in training, yet preserve redundant reasoning after the policy becomes more competent. CARE addresses this problem by evolving the reward preference over reasoning effort according to competence, encouraging exploration when competence is limited and progressively favoring concise, information-dense reasoning as competence improves.}
\label{fig:motivation}
\end{figure*}

\section{Introduction}

Large multimodal language models (MLLMs) have recently shown strong performance on reasoning tasks that require jointly processing heterogeneous inputs such as images, documents, charts, and videos \cite{zhang2023multimodalcot,wang2025mcotsurvey}. Among these settings, video question answering (VideoQA) is particularly demanding because successful prediction depends not only on recognizing visual content, but also on tracking temporal changes, associating evidence across frames, and composing multiple intermediate inferences over a dynamic visual stream. To improve reliability in such compositional problems, recent post-training methods increasingly encourage models to produce explicit chain-of-thought (CoT) rationales before giving the final answer. This shifts inference from direct answer prediction to an explicit reasoning process, and has led to substantial gains in multimodal reasoning performance \cite{zhang2023multimodalcot,feng2025videor1}.

Once explicit reasoning is introduced, however, post-training no longer optimizes answer correctness alone. It also shapes how much reasoning computation the model is encouraged to spend before committing to an answer. In this sense, reasoning length becomes a training-time control variable rather than a superficial property of the output. Recent studies on reasoning models have shown that longer reasoning does not monotonically produce better solutions: models may overthink simple instances with unnecessarily verbose derivations, while still underthinking difficult ones by failing to allocate sufficient computation when additional reasoning is required \cite{su2025between,sui2025stop}. These observations indicate that reasoning length should not be treated as a fixed proxy for reasoning quality. Instead, it should be viewed as a controllable reasoning budget whose desirable allocation depends on both model competence and instance difficulty \cite{su2025adaptive,liu2025laser}.

This issue is more consequential in multimodal reasoning, where the usefulness of extended reasoning must be balanced against the need to remain grounded in perceptual evidence. In vision-language settings, long reasoning traces can increase the risk of drifting away from the underlying visual signal, thereby amplifying hallucinations and over-reliance on language priors \cite{yu2026modality,ghosh2024vdgd,yu2025unicorn,thinkinglessseeing2025}. For video reasoning, this tension is even sharper because the model must integrate temporally distributed evidence while maintaining consistency over a longer observation horizon. Recent reinforcement learning (RL) methods have demonstrated that RL can improve temporal and compositional reasoning in MLLMs \cite{feng2025videor1}, but they still leave a central question insufficiently addressed: as the policy becomes more capable during training, how should the preferred amount of reasoning change?

Existing approaches usually regulate reasoning length with static control strategies, such as fixed rewards, penalties, or thresholds. Although these designs can suppress obvious verbosity, they also assume that the preferred reasoning budget should remain unchanged throughout training. This assumption is difficult to justify in RL, where model competence evolves continuously. Early in training, when reasoning ability is still limited, longer trajectories can be useful because they enlarge the space of candidate inference paths, expose more intermediate states, and support exploration. Later in training, once the policy has acquired stronger reasoning ability, the same preference for long trajectories can instead preserve redundant narration, repeated self-verification, or padding behaviors that consume tokens without improving the solution. The key problem is therefore not merely excessive length or insufficient brevity, but a competence--constraint mismatch: static reward design imposes a fixed preference on a behavior whose utility changes as the policy improves.

As summarized in Fig.~\ref{fig:motivation}, this mismatch and the intuition that follows from it highlight how fixed length preferences can create two opposite optimization failures during training. If the reward favors brevity too early, the policy may prematurely truncate trajectories that are still useful for exploration and error correction. If the reward continues to tolerate or encourage long reasoning after competence has improved, the policy may retain redundant reasoning that contributes little beyond token overhead. The same output length therefore has different implications at different stages of learning. More broadly, static designs judge reasoning effort in isolation, without accounting for whether the current behavior arises from a weak or strong policy, whether the instance is easy or difficult, or whether a successful long trajectory is merely routine or genuinely informative under the current competence level.

These observations suggest that reasoning-length control should be formulated as competence-aware reasoning budget allocation rather than static regularization. Based on this view, we propose CARE, short for Competence-Aware Reward Shaping, a training-time reward framework that adapts how reasoning effort is valued during RL for multimodal video reasoning. CARE maintains a smoothed estimate of model competence, uses this estimate to route optimization into progressively different behavioral regimes, calibrates reasoning effort relative to batch context instead of raw length alone, and further amplifies behaviors that are unexpectedly successful under the current competence level. In this way, CARE does not rely on a separate inference-time controller. Instead, it reshapes the reward landscape during training so that an adaptive reasoning policy emerges naturally: more reasoning is preserved when difficult cases still require it, while easier cases are gradually compressed into more concise and information-dense solutions.

We summarize the main contributions of this work as follows:

\begin{enumerate}
\item We identify a competence--constraint mismatch in multimodal reasoning RL: static reasoning-length rewards implicitly impose a fixed reasoning-budget preference throughout training, despite the fact that the desirable budget is inherently non-stationary as model competence evolves.

\item We propose CARE, a competence-aware reward shaping framework that progressively adjusts reasoning incentives from exploration-oriented long-form reasoning to efficiency-oriented concise reasoning, implemented through competence-based stage routing, difficulty-aware modulation, and posterior reward amplification.

\item We show that CARE induces an emergent competence-aware reasoning policy without any explicit inference-time controller: the trained model self-organizes to allocate more reasoning to hard instances and less to easy ones, yielding improved accuracy, lower token consumption, and more stable training dynamics across multiple video reasoning benchmarks.
\end{enumerate}

\begin{figure*}[t]
\centering
\includegraphics[width=\textwidth]{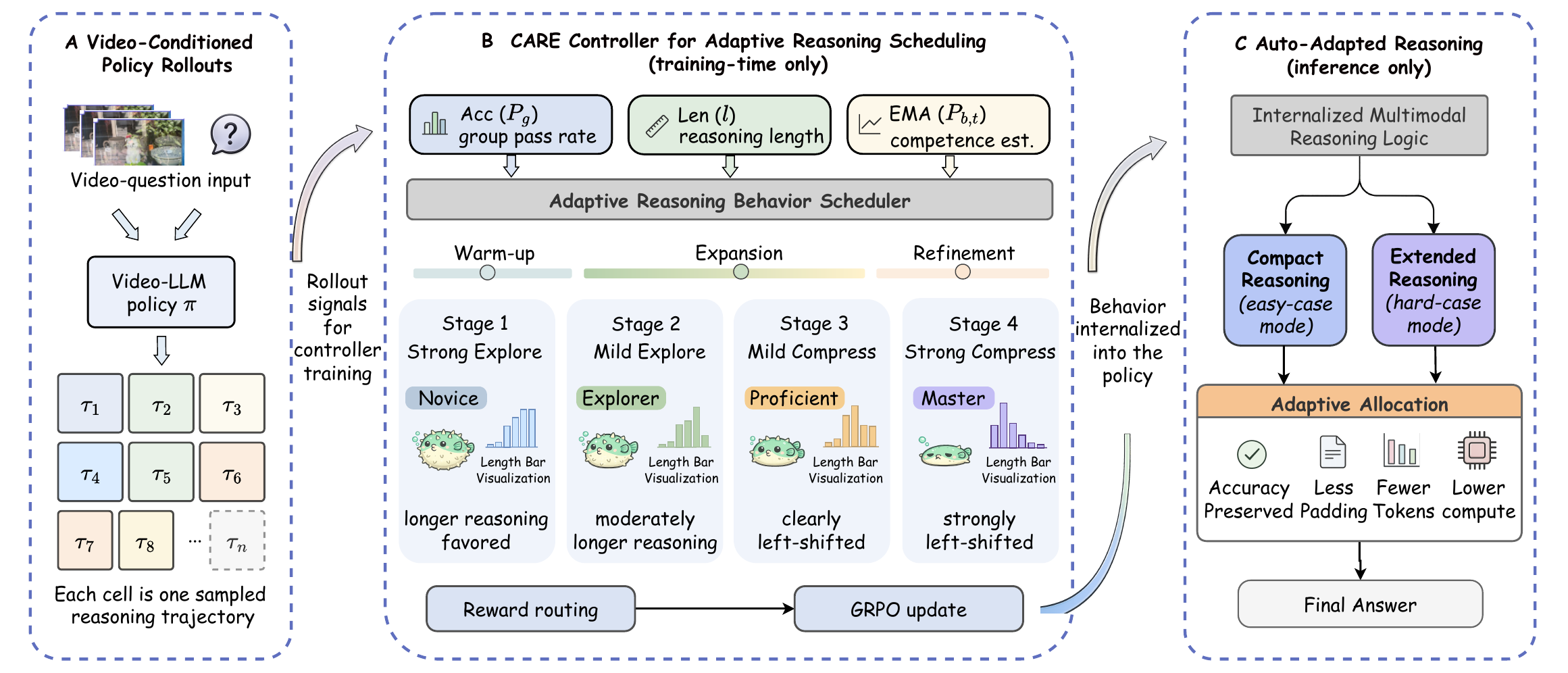}
\caption{Overview of CARE. (A) The video-conditioned policy rollouts produce reasoning trajectories. (B) The CARE controller dynamically routes reward incentives according to competence estimates during reinforcement learning. (C) This training-time regulation internalizes the adaptive reasoning behavior, enabling auto-adapted allocation of compact or extended reasoning during inference. Overall, this framework establishes a principled balance between reasoning exploration and execution efficiency.}
\label{fig:framework}
\end{figure*}

\section{related work}
\subsection{RL for Video Reasoning in MLLMs}

Recent progress in multimodal post-training has increasingly extended reinforcement learning from text and image reasoning to the video domain. A representative line of work shows that RL can substantially strengthen video reasoning in MLLMs by improving temporal understanding, compositional inference, and generalization beyond supervised chain-of-thought distillation alone \cite{feng2025videor1,VIDEORFT}. In particular, Video-R1 is among the first frameworks to systematically adapt the R1-style reinforcement paradigm to video reasoning, highlighting two central challenges in this setting: temporal modeling and the scarcity of high-quality video reasoning data \cite{feng2025videor1}. VideoRFT further demonstrates that reinforcement fine-tuning can effectively incentivize human-like video reasoning when combined with scalable construction of video CoT data \cite{VIDEORFT}.

Beyond the video domain specifically, recent multimodal reinforcement fine-tuning methods also provide important foundations for video reasoning. Visual-RFT shows that rule-based verifiable rewards can effectively transfer the RFT paradigm to visual perception and multimodal reasoning tasks \cite{liu2025visualrft}, while Reason-RFT demonstrates that GRPO-based reinforcement learning improves the generalization and data efficiency of visual reasoning models compared with purely supervised CoT training \cite{tan2025reasonrft}. Point-RFT further suggests that visually grounded CoT is more effective than text-only reasoning for multimodal reasoning, underscoring the importance of grounding the reasoning process in visual evidence rather than treating reasoning as purely linguistic elaboration \cite{ni2025pointrft}. These developments collectively indicate that RL is becoming a practical paradigm for strengthening multimodal reasoning, especially when reward design and reasoning traces are made compatible with visual evidence.

At the same time, the video setting introduces distinctive challenges that make reasoning control more difficult than in static-image or text-only tasks. Benchmarks such as V-STaR show that video reasoning requires jointly modeling what objects are present, when events occur, and where they are located, making spatio-temporal reasoning a first-class challenge rather than a byproduct of general video understanding \cite{cheng2025vstar}. Existing RL-based video reasoning methods therefore mainly focus on temporal-aware optimization, reasoning data construction, or grounding-oriented reward design \cite{feng2025videor1,VIDEORFT}. However, they pay less attention to another important dimension of post-training: how much reasoning budget the model should be encouraged to spend at different stages of learning. In other words, while prior work has substantially improved the reasoning content of video MLLMs, it has rarely modeled how the preferred amount of reasoning should evolve as the model becomes more competent. Our work addresses this gap by formulating reasoning-length regulation as a competence-aware and non-stationary budget allocation problem during RL.

\subsection{Reasoning Efficiency and Budget Allocation}

As reasoning models become stronger, improving reasoning efficiency has emerged as an important research problem rather than a secondary deployment concern. Recent studies show that longer reasoning does not monotonically lead to better performance: models may overthink simple problems by generating unnecessarily verbose chains, yet underthink harder ones by failing to extend reasoning when additional computation is needed \cite{su2025between,sui2025stop}. This observation has motivated a growing body of work on efficient reasoning, including response compression, shorter reasoning distillation, dynamic early stopping, and reward-based length control \cite{sui2025stop}. A first line of work improves efficiency mainly at inference time through post-hoc control over generated reasoning. These methods reduce redundant computation by compressing intermediate thoughts, terminating reasoning early, or encouraging shorter outputs after reasoning capabilities have already been acquired. While such strategies are effective for lowering inference cost, they primarily operate on completed or partially generated trajectories, and therefore treat reasoning length as an output-side phenomenon to be corrected after the reasoning policy has already been learned \cite{sui2025stop}.

A second line of work brings efficiency considerations into training by explicitly shaping rewards with respect to reasoning length. For example, a method proposes adaptive rewards to better balance answer accuracy and reasoning length during optimization \cite{su2025adaptive}, while LASER formulates efficient reasoning methods under a unified length-based reward shaping framework and uses target-length-controlled step rewards to improve the Pareto trade-off between performance and efficiency \cite{liu2025laser}. These methods move beyond purely post-hoc adjustment by recognizing that reasoning efficiency should be learned during training rather than corrected only at inference time.

Despite this progress, existing training-time approaches still typically regulate reasoning efficiency through fixed targets, local heuristics, or sample-level adjustments. As a result, they do not explicitly model a key property of reinforcement learning for reasoning: the preferred amount of reasoning is itself non-stationary during training. A reasoning budget that is beneficial when the model is still exploring candidate solution paths may become wasteful once the model has acquired stronger reasoning competence. In this sense, the core challenge is not merely to shorten reasoning, but to adapt the reward preference over reasoning length as the model evolves. Our work addresses this gap by framing reasoning efficiency as a competence-aware and non-stationary budget allocation problem in multimodal RL training.

\begin{figure*}[!t]
\centering
\includegraphics[width=\textwidth]{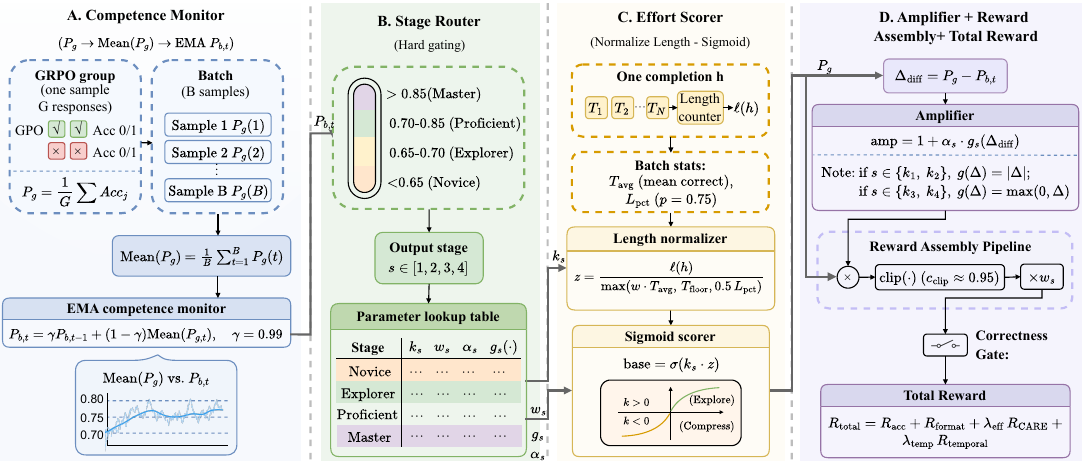}
\caption{Pipeline of the proposed CARE algorithm.}
\label{fig:algo}
\end{figure*}

\section{Method}

CARE is a training-time reward controller integrated into the GRPO optimization loop. Rather than applying a static preference for either long or short chains of thought, CARE adjusts the incentive on reasoning length according to the current competence of the policy. As illustrated in Fig.~\ref{fig:framework}, the framework follows a sequential pipeline: it first observes group-level rollout outcomes, then derives a stable competence signal, converts that signal into a stage-aware effort score, and finally assembles the reward used for policy optimization.

\subsection{Overview of CARE}

For each prompt, the policy produces a rollout group and yields three training-time signals: the group pass rate $P_g$, the reasoning length $\ell(h)$, and the EMA-smoothed competence estimate $P_{b,t}$. CARE uses these signals in a fixed order. It first maps the competence estimate to a training stage $s$, then normalizes the reasoning length into a batch-relative effort score $z$, and subsequently modulates the effort preference according to both the current stage and the informativeness of the observed success. The resulting reward is therefore conditioned jointly on policy maturity, local difficulty, and posterior surprise, instead of relying on a single global length heuristic.

These intermediate signals are then integrated to formulate the final CARE reward as follows:
\begin{equation}
R_{CARE} = R_{acc}\cdot w_s \cdot \min\!\big(\eta, \sigma(k_s z)\cdot A\big),
\end{equation}
where $k_s$ controls the stage-aware effort preference, $A$ is a posterior amplification factor, $w_s$ is a stage-dependent weight, and $\eta$ is a clipping threshold. 

\subsection{Competence Estimation}

The reward controller requires a stable estimate of model maturity. Using instantaneous batch accuracy directly would be problematic because batch difficulty varies substantially during training, which would cause frequent oscillation in the routing signal. We therefore construct a historical competence estimate through an exponential moving average (EMA), allowing the reward policy to track long-term capability trends rather than transient batch noise.

For each question, the policy generates $G$ rollouts in a GRPO group. We compute the group pass rate as follows:
\begin{equation}
P_g = \frac{1}{G} \sum_{i=1}^{G} Acc_i ,
\end{equation}
where $Acc_i \in \{0,1\}$ indicates whether the $i$-th rollout is correct.
Given a batch containing $B$ groups, we then compute the average batch pass rate as follows:
\begin{equation}
\mathrm{Mean}(P_g) = \frac{1}{B} \sum_{t=1}^{B} P_g(t).
\end{equation}
The competence estimate is updated according to the following equation:
\begin{equation}
P_{b,t} = \gamma P_{b,t-1} + (1-\gamma)\mathrm{Mean}(P_g),
\end{equation}
where $\gamma$ is the smoothing coefficient.
In our implementation, we set $\gamma=0.99$ so that routing decisions evolve smoothly with training. To avoid initialization bias, the EMA is warm-started from the observed pass rate at the beginning of training rather than from an arbitrary constant. As a result, $P_{b,t}$ serves as a stable global state variable for the subsequent reward controller.

\subsection{Stage-Aware Effort Scoring}

The effect of reasoning length depends on the current competence regime. When the policy is still weak, longer chains may reflect necessary exploration; when the policy is already competent, excessive length more often indicates redundancy. CARE therefore does not assign a fixed preference to long outputs. Instead, it first routes training into a competence stage and then evaluates length relative to the local batch context.

The current stage is determined as follows:
\begin{equation}
s = f(P_{b,t}), \quad s \in \{\text{Novice}, \text{Explorer}, \text{Proficient}, \text{Master}\}.
\end{equation}
Specifically, the training process is divided into four progressive stages based on a set of predefined competence thresholds $\mathcal{T}$. In the Novice stage, the model is still weak and often fails to complete complex reasoning correctly, prompting CARE to strongly protect long-form exploration. As the model advances to the Explorer stage, it begins to exhibit useful reasoning patterns but still benefits from additional exploration; thus, CARE continues to encourage longer reasoning, albeit more moderately. Upon reaching a more reliable competence level in the Proficient stage, blindly encouraging longer reasoning is no longer desirable, and the controller begins to shift toward efficiency. Finally, in the Master stage, the model is expected to reason accurately and efficiently, leading CARE to strongly penalize redundant reasoning tokens. The exact values of $\mathcal{T}$ are detailed in the experimental setup. These thresholds postpone compression until the policy has moved sufficiently beyond its SFT initialization, thereby reducing the risk of discouraging useful exploration too early.

Raw response length is not directly comparable across instances, because difficult samples naturally require larger reasoning budgets. We therefore measure effort only after batch-relative normalization. Let $\ell(h)$ denote the token length of the generated reasoning chain. The normalized effort score is calculated as follows:
\begin{equation}
z = \frac{\ell(h)}{\max(\omega \cdot T_{\mathrm{avg}}, T_{\mathrm{floor}}, 0.5 L_{\mathrm{pct}})},
\end{equation}
where $T_{\mathrm{avg}}$ is the average length of correct responses within the batch, $L_{\mathrm{pct}}$ denotes the $75^{\text{th}}$ percentile of response lengths, $T_{\mathrm{floor}}$ is a fixed lower bound for numerical stabilization, and $\omega$ is a tolerance multiplier that relaxes the length penalty to prevent premature truncation.
The denominator combines successful-response statistics with a conservative floor so that normalization remains meaningful even when correct samples are sparse or abnormally short. In early training, this prevents the effort baseline from collapsing together with model performance.

The stage-aware preference is then encoded through a modulated coefficient $k_s$. We first define a stage-specific base routing coefficient $b_s$, and then adjust it according to both group difficulty and the model's distance from the competence anchor:
\begin{equation}
k_s = b_s + \mathrm{sgn}(b_s) \big( \alpha (1 - P_g) + \beta_{\mathrm{mod}} |P_{b,t} - a| \big),
\end{equation}
where $b_s \in \mathcal{B}$ is the base routing coefficient for stage $s$, $a$ is the competence anchor, and $\alpha, \beta_{\mathrm{mod}}$ are modulation weights. Positive $k_s$ values encourage longer reasoning, whereas negative values penalize it. Larger magnitudes are used when the training objective should be more explicit, while smaller magnitudes are used near transition regions to avoid abrupt changes in the reward landscape.

Based on $z$ and $k_s$, we compute the stage-aware effort score as follows:
\begin{equation}
S = \sigma(k_s z),
\end{equation}
which quantifies whether the observed effort is desirable under the current competence regime.

\subsection{Posterior Amplifier}

Stage-aware effort scoring specifies the preferred direction of reasoning behavior, but it does not distinguish routine outcomes from especially informative ones. To address this, CARE adds a posterior amplifier that measures how far the current group performance exceeds or deviates from the expected competence baseline.

We first compute the relative advantage as follows:
\begin{equation}
\Delta_{diff} = P_g - P_{b,t}.
\end{equation}
A larger positive value indicates that the current group performs better than what the historical competence estimate would predict. The amplification factor is then defined as follows:
\begin{equation}
A = 1 + a_s g_s(\Delta_{diff}),
\end{equation}
where $a_s \in \mathcal{A}_{\mathrm{amp}}$ is a stage-specific amplifier weight and $g_s(\cdot)$ is a stage-dependent transformation.
We use different transformations in different stages:
\begin{equation}
g_s(\Delta) =
\begin{cases}
|\Delta|, & s \in \{\text{Novice}, \text{Explorer}\},\\
\max(0,\Delta), & s \in \{\text{Proficient}, \text{Master}\}.
\end{cases}
\end{equation}
In early stages, the goal is to remain sensitive to unusual and potentially useful exploratory behavior, even when it is not yet consistently aligned with the historical average. In later stages, the controller becomes more selective and amplifies only outcomes that outperform expectations. This stage-dependent asymmetry allows the amplifier to function as an auxiliary booster without overriding the main effort preference established by the stage-aware scorer.

\subsection{Final Reward and Training Objective}

The raw CARE effort reward is assembled as follows:
\begin{equation}
R_{CARE\_raw} = w_s \cdot \min(\eta, S \cdot A),
\end{equation}
where $w_s \in \mathcal{W}$ is a stage-dependent weight and $\eta$ is a clipping threshold applied to prevent reward saturation.
Here, $S$ scores the desirability of the current reasoning effort, $A$ scales that score according to posterior informativeness, and $w_s$ adjusts the influence of the effort term across training stages. In practice, larger stage weights are assigned when exploration should be emphasized, whereas smaller weights are used when the effort term should act mainly as an efficiency regularizer.

Reasoning effort should not be rewarded independently of factual correctness. We therefore gate the effort term by the accuracy reward, formulated as follows:
\begin{equation}
R_{CARE} = R_{CARE\_raw} \cdot R_{acc}.
\end{equation}
The final reinforcement learning objective is defined as follows:
\begin{equation}
R_{total}
=
R_{acc}
+
R_{format}
+
R_{CARE}
+
\lambda_{\mathrm{temp}}R_{temporal},
\end{equation}
where $R_{acc}$ is the task accuracy reward, $R_{format}$ is the formatting reward, $R_{CARE}$ is the CARE effort reward, and $R_{temporal}$ serves as an auxiliary temporal reward where applicable.
Clipping is applied after reward assembly to prevent a small number of amplified samples from dominating the GRPO update. All CARE components operate only during training, so the method reshapes reasoning behavior without introducing additional inference-time computation.

\section{Experiments}

\subsection{Settings}

We first conduct one epoch of supervised fine-tuning on Video-R1-CoT-165k to initialize the reasoning capability of the model, and then continue training with CARE under the GRPO framework. Unless otherwise specified, all experiments are conducted on 8 NVIDIA L20 (46GB) GPUs with a per-device batch size of 1. For reinforcement learning, we use Video-R1-260k, which contains about 260K video-question-answer pairs with visual reasoning queries. During training, each sample is limited to at most 16 frames with dynamic resolution up to $128 \times 28 \times 28$. During inference, the resolution is increased to $256 \times 28 \times 28$, and the number of frames is extended to 16$\sim$64. We set the GRPO group size to $G=8$ and train for 5 epochs.

Table~\ref{tab:settings} summarizes the optimization and CARE-specific hyperparameters used throughout training and evaluation. Unless otherwise specified, stage-dependent parameter sets are listed in the order of Novice, Explorer, Proficient, and Master.

\begin{table}[t]
  \caption{Optimization and CARE hyperparameters.}
  \label{tab:settings}
  \begin{center}
    \begin{footnotesize}
      \begin{sc}
        \setlength{\tabcolsep}{3pt}
        \renewcommand{\arraystretch}{1.08}
          \begin{tabular}{>{\raggedright\arraybackslash}p{0.54\columnwidth}>{\raggedright\arraybackslash}p{0.34\columnwidth}}
            \toprule
            Hyperparameter & Value \\
            \midrule
            \multicolumn{2}{c}{Optimization} \\
            \midrule
            Optimizer & AdamW \\
            Learning rate & $1\times10^{-6}$ \\
            Weight decay & 0.01 \\
            KL coefficient & 0.04 \\
            Gradient clipping & 5.0 \\
            \midrule
            \multicolumn{2}{c}{CARE Controller} \\
            \midrule
            EMA momentum ($\gamma$) & 0.99 \\
            Competence thresholds ($\mathcal{T}$) & $\{0.65, 0.70, 0.75\}$ \\
            Competence anchor ($a$) & 0.70 \\
            Base routing coefficients ($\mathcal{B}$) & $\{+3.0, +1.2, -1.2, -3.0\}$ \\
            Modulation weight ($\alpha$) & 0.5 \\
            Boundary modulation ($\beta_{\mathrm{mod}}$) & 0.3 \\
            Stage weights ($\mathcal{W}$) & $\{0.3, 0.25, 0.2, 0.2\}$ \\
            Amplifier weights ($\mathcal{A}_{\mathrm{amp}}$) & $\{1.0, 0.5, 0.2, 0.5\}$ \\
            Length tolerance ($\omega$) & 2.5 \\
            Length floor ($T_{\mathrm{floor}}$) & 300 \\
            Reward clip ($\eta$) & 0.95 \\
            Temporal coefficient ($\lambda_{\mathrm{temp}}$) & 1.0 \\
            \bottomrule
          \end{tabular}
      \end{sc}
    \end{footnotesize}
  \end{center}
  \vskip 0.1in
\end{table}

\subsection{Main Results}
\begin{table*}[t]
  \caption{Video model performance comparison.}
  \label{tab:main_results}
  \begin{center}
    \begin{small}
      \begin{sc}
      \resizebox{\textwidth}{!}{
            \begin{tabular}{lcccc|ccc}
              \toprule
              && \multicolumn{3}{c}{Video Reasoning Benchmark} & \multicolumn{3}{c}{Video General Benchmark} \\
              \cmidrule(r){3-5} \cmidrule(l){6-8}
              Models & Frames & VSI-Bench & VideoMMMU & MMVU(mc) & MVBench & TempCompass & VideoMME(wo sub) \\
                    
              \midrule
              LLaMA-VID \cite{Llama-vid} & - & - & - & - & 41.9 & 45.6 & - \\
              VideoLLaMA2 \cite{VIDEOLLAMA2} & - & - & - & 44.8 & 54.6 & - & 47.9 \\
              LongVA-7B \cite{LONGVA} & - & 29.2 & 23.9 & - & - & 56.9 & 52.6 \\
              VILA-1.5-8B \cite{VILA-1.5} & - & 28.9 & 20.8 & - & - & 58.8 & - \\
              VILA-1.5-40B \cite{VILA-1.5} & - & 31.2 & 34.0 & - & - & - & 60.1 \\
              Video-UTR-7B \cite{VIDEO-UTR} & - & - & - & - & 58.8 & 59.7 & 52.6 \\
              LLaVA-OneVision-7B \cite{LLAVA-ONEVISION} & - & 32.4 & 33.8 & 49.2 & 56.7 & - & 58.2 \\
              Kangeroo-8B \cite{KANGEROO} & - & - & - & - & 61.1 & 62.5 & 56.0 \\
              Qwen2.5-VL-7B \cite{Qwen2.5-VL} & - & - & 47.4 & 61.3 & 59.4 & 69.2 & 52.8 \\
              \midrule
              Video-R1-7B \cite{feng2025videor1} & 16 & 30.3 & 47.2 & 63.5 & 62.4 & 70.8 & 54.3 \\
              DeepVideo-R1 \cite{park2025deepvideo} & - & 33.0 & 40.7 & 59.0 & 49.6 & 63.1 & 51.1 \\
              TinyLLaVA-Video-R1 \cite{TinyLLaVA-Video-R1} & 16 & - & - & 46.9 & - & 49.5 & 46.6 \\
              VIDEORFT \cite{VIDEORFT} & 32 & - & - & 51.1 & 62.1 & - & - \\
              Temporal-RLT \cite{Temporal-RLT} & 32 & - & - & 65.0 & - & - & 57.6 \\
              Video-COM \cite{rasheed2025video}    & - & - & 50.2&65.4&-&71.3&-\\
              \midrule
              \rowcolor{ourscolor} Ours & 16 & 33.9 & 50.2 & 64.2 & 64.4 & 73.1 & 57.3 \\
              \rowcolor{ourscolor} Ours & 32 & 34.3 & 51.0 & 64.5 & 65.0 & 73.6 & 60.6 \\
              \rowcolor{ourscolor} Ours & 64 & \textbf{36.2} & \textbf{51.1} & \textbf{64.8} & \textbf{65.7} & \textbf{73.8} & \textbf{62.3} \\
              \bottomrule
            \end{tabular}
        }
      \end{sc}
    \end{small}
  \end{center}
  \vskip 0.2in
\end{table*}

To comprehensively evaluate the model's spatial-temporal perception and complex reasoning capabilities, we conduct experiments on a suite of video reasoning benchmarks: VSI-Bench \cite{yang2025thinking}, VideoMMMU \cite{hu2025video}, and MMVU \cite{zhao2025mmvu}. Additionally, we assess general video understanding capabilities using MVBench \cite{li2024mvbench}, TempCompass \cite{liu2024tempcompass}, and Video-MME \cite{fu2025video}.

Table~\ref{tab:main_results} compares our method with recent video-language models and reinforcement-learning-based video reasoning approaches.

\paragraph{Overall Performance and Reasoning Capabilities}
Our method consistently outperforms previous models across the evaluated benchmarks. Under the identical Qwen2.5-VL-7B backbone and 16-frame setting, CARE achieves notable gains over the Video-R1 baseline on reasoning-intensive datasets, improving VSI-Bench from 30.3 to 33.9 and VideoMMMU from 47.2 to 50.2. These results demonstrate that competence-aware reward shaping effectively enhances multi-step video reasoning.

\paragraph{General Video Understanding and Scaling}
Beyond specialized reasoning tasks, CARE improves general video understanding, yielding higher scores on MVBench (64.4 vs.\ 62.4), TempCompass (73.1 vs.\ 70.8), and VideoMME (57.3 vs.\ 54.3). Furthermore, scaling the input from 16 to 64 frames provides consistent performance gains, reaching 36.2 on VSI-Bench and 62.3 on VideoMME, indicating that CARE robustly leverages richer temporal contexts while maintaining stable reasoning behavior.

\subsection{Ablation Study}

To quantify the contribution of each component in CARE, we conduct ablation studies by removing or simplifying key designs, as shown in Table~\ref{tab:ablation}.

\begin{table*}[t]
  \caption{Ablation study of CARE.}
  \label{tab:ablation}
  \begin{center}
    \begin{small}
      \begin{sc}
      \resizebox{\textwidth}{!}{
            \begin{tabular}{lcccc|ccc}
              \toprule
              && \multicolumn{3}{c}{Video Reasoning Benchmark} & \multicolumn{3}{c}{Video General Benchmark} \\
              \cmidrule(r){3-5} \cmidrule(l){6-8}
              Models & Frames & VSI-Bench & VideoMMMU & MMVU(mc) & MVBench & TempCompass & VideoMME(wo sub) \\
              \midrule
              Baseline & 16 & 30.3 & 47.2 & 63.5 & 62.4 & 70.8 & 54.3 \\
              Ours (2-stage)  & 16 & 29.4 & 48.3 & 63.8 & 63.2 & 71.1 & 55.2 \\
              Ours (SFT-only) & 16 & 30.2 & 44.6 & 59.2 & 57.1 & 69.4 & 51.9 \\
              Ours (RL-only)  & 16 & 31.2 & 46.8 & 63.2 & 60.8 & 71.1 & 53.4 \\
              Ours w/o Amplifier & 16 & 30.7 & 48.8 & 63.9 & 63.4 & 70.8 & 54.1 \\
              Ours w/o EMA    & 16 & 30.4 & 47.5 & 61.6 & 62.2 & 71.0 & 53.0 \\
              Ours (Fixed-step) & 16 & 30.7 & 44.1 & 61.4 & 54.3 & 69.1 & 50.2 \\
              Ours w/o CARE   & 16 & 32.7 & 48.3 & 62.1 & 61.1 & 71.3 & 54.5 \\
              \midrule
              \rowcolor{ourscolor} Ours & 16 & 33.9 & 50.2 & 64.2 & 64.4 & 73.1 & 57.3 \\
              \bottomrule
            \end{tabular}
        }
      \end{sc}
    \end{small}
  \end{center}
  \vskip 0.2in
\end{table*}

\begin{figure}[!t]
\centering
\includegraphics[width=\columnwidth]{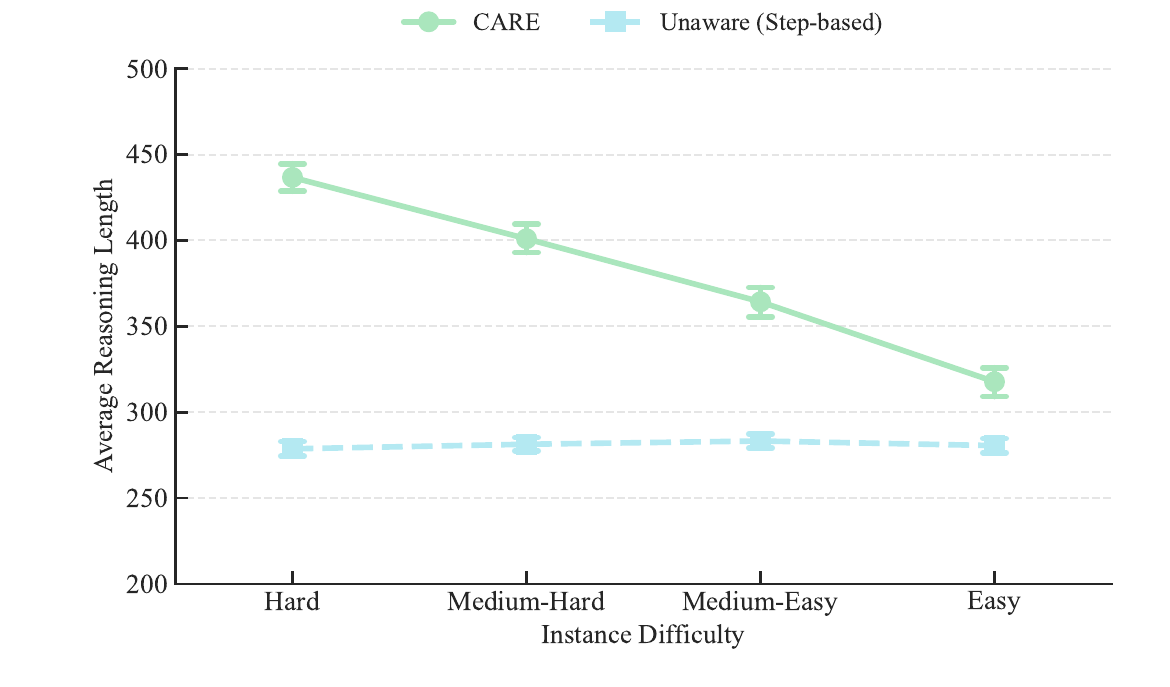}
\caption{Visualization of the fixed-step ablation in Table~\ref{tab:ablation}. In the Ours (Fixed-step) setting, the resulting policy allocates a nearly constant reasoning length across different instance difficulty levels (categorized by the pass rate). In contrast, CARE dynamically adjusts the reasoning budget, preserving a clear monotonic allocation pattern from hard to easy instances.}
\label{fig:step_based_curve}
\end{figure}

\paragraph{Effect of Fixed-step Scheduling}
As illustrated in Fig.~\ref{fig:step_based_curve}, this ablation evaluates reasoning-length allocation by comparing CARE with the Ours (Fixed-step) variant reported in Table~\ref{tab:ablation}. Ours (Fixed-step) replaces the competence-aware transitions with a fixed schedule, evenly dividing the training process into four stages (Novice, Explorer, Proficient, and Master) that each occupy one quarter of the total training steps. In this setting, the reward controller is unaware of the instance-level competence signals (such as the group pass rate $P_g$) and enforces stage transitions based solely on step boundaries rather than sample-adaptive control. Consequently, the resulting policy maintains a nearly constant reasoning length across different instance difficulty levels (categorized by the pass rate). In contrast, CARE allocates substantially longer reasoning to hard instances and progressively compresses reasoning as difficulty decreases. Furthermore, this variant remains consistently below CARE on all evaluated benchmarks, reducing VideoMMMU from 50.2 to 44.1, MVBench from 64.4 to 54.3, and VideoMME from 57.3 to 50.2. This indicates that equal-duration stage allocation is insufficient. Without sample-adaptive control, the model cannot adjust the reasoning budget according to the actual competence-demand mismatch of different instances, leading to less effective reasoning allocation and weaker overall performance.

\paragraph{Effect of Interval Granularity}
Replacing the four-stage competence partition with a binary interval (Ours (2-stage)) significantly degrades performance, dropping VSI-Bench from 33.9 to 29.4. A coarse two-stage reward fails to capture the gradual progression of model competence, introducing overly abrupt reward signals that disrupt the smooth transition from exploration to compression. This confirms the necessity of fine-grained competence modeling.

\paragraph{Effect of Posterior Amplification and $T_{\text{floor}}$}
Removing the posterior amplifier (Ours w/o Amplifier) causes consistent performance drops, with VideoMMMU decreasing from 50.2 to 48.8. This indicates that base length control alone is insufficient, while assigning larger rewards to unexpectedly strong performances accelerates learning on difficult instances. Furthermore, removing the minimum reasoning length constraint ($T_{\text{floor}}$) leads to training collapse, where the model rapidly converges to extremely short, trivial responses. $T_{\text{floor}}$ is thus critical to ensure sufficient capacity for multi-step reasoning.

\paragraph{Effect of EMA Smoothing}
Using the instantaneous batch mean pass rate instead of the EMA competence monitor (Ours w/o EMA) causes clear degradation. Because video RL typically uses small batch sizes, raw pass rates are highly sensitive to sample difficulty, leading to severe reward shocks that can alternate abruptly between length penalization and encouragement. EMA smoothing provides a stable competence tracker, ensuring continuous curriculum progression and stable policy learning.

\paragraph{Effect of Cold-start SFT}
Ours (SFT-only) only applies cold-start supervised fine-tuning to the base model and yields the weakest overall performance, with 30.2 on VSI-Bench, 44.6 on VideoMMMU, and 57.1 on MVBench. This result indicates that cold-start SFT improves initialization but does not provide sufficient optimization pressure for video reasoning and temporal understanding.

\paragraph{Effect of Removing CARE}
Ours w/o CARE preserves the same training pipeline but removes CARE-specific adaptive reward design. Its performance remains consistently below the full method, for example dropping VSI-Bench from 33.9 to 32.7 and VideoMMMU from 50.2 to 48.3. This result indicates that the gain does not come from the training recipe alone. Instead, CARE's competence-aware reward shaping provides the key optimization signal that converts reinforcement learning into stronger and more stable improvements.

\paragraph{Effect of Zero-start RL}
Ours (RL-only) directly applies standard GRPO to the base model without cold-start supervised fine-tuning. It consistently outperforms Ours (SFT-only) on most benchmarks, improving VSI-Bench from 30.2 to 31.2 and MVBench from 57.1 to 60.8, but still remains below both Ours w/o CARE and the full method across all datasets. This result shows that generic RL optimization is beneficial, but starting RL without cold-start initialization still leaves the policy at a clear disadvantage. CARE further improves upon this foundation by providing competence-aware reward shaping, translating policy optimization into more stable and consistent gains on both video reasoning and general video benchmarks.

\subsection{Analysis of Static Length Normalization}
\begin{figure}[t]
    \centering
    \includegraphics[width=\linewidth]{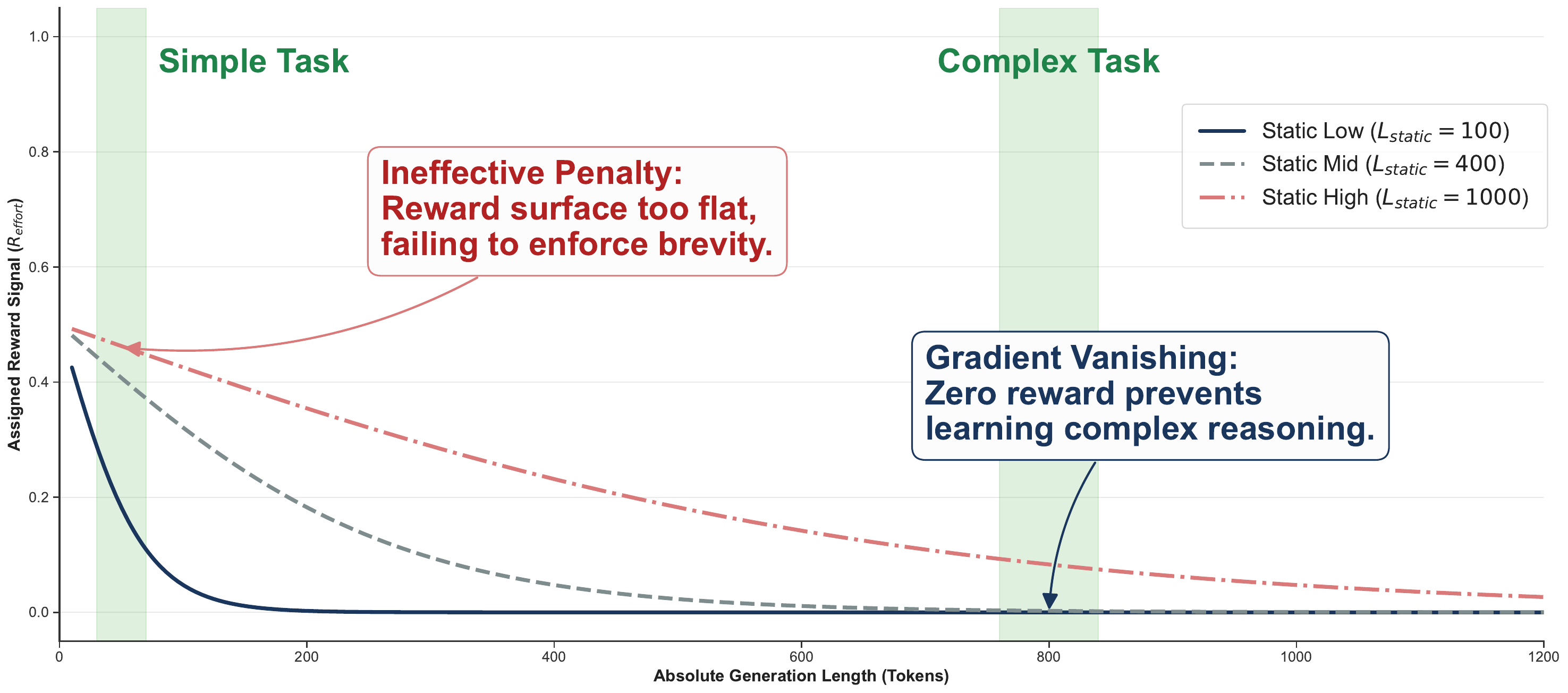}
    \caption{Optimization landscape of static length normalization under different fixed denominators $L_{\text{static}}$. A single static denominator cannot accommodate varying reasoning lengths across instances.}
    \label{fig:static_length}
\end{figure}

We further analyze why static length normalization is fundamentally unsuitable for multimodal reasoning. 
As shown in Fig.~\ref{fig:static_length}, using a fixed denominator $L_{\text{static}}$ produces incompatible optimization behavior across tasks with different intrinsic reasoning demands. 
When $L_{\text{static}}$ is small, long yet necessary reasoning chains are pushed into the saturated tail of the reward function, yielding near-zero gradients and encouraging premature truncation. 
When $L_{\text{static}}$ is large, the reward surface becomes too flat in the short-length region, making it difficult to penalize redundant reasoning on simple instances. 
Intermediate values merely trade off these two failure modes rather than resolving them. 
This analysis suggests that length calibration should be adaptive to task difficulty, rather than enforced through a single global normalization constant.

\subsection{Training Behavior During RL}

\begin{figure*}[t]
\centering
\includegraphics[width=\textwidth]{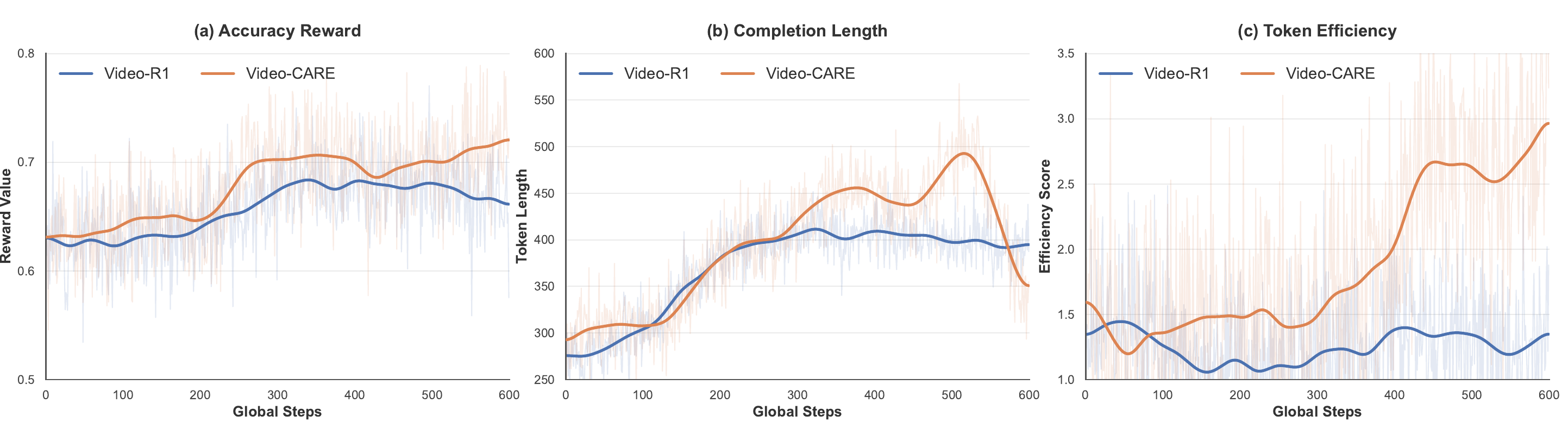}
\caption{Training behavior during reinforcement learning. Left: accuracy reward. Middle: reasoning length. Right: token efficiency. Compared with Video-R1, CARE attains higher reward, adjusts response length more adaptively, and achieves better token efficiency throughout training.}
\label{fig:db_3t1}
\end{figure*}

As summarized in Fig.~\ref{fig:db_3t1}, the training dynamics of Video-R1 and CARE are evaluated from three complementary perspectives: accuracy reward, reasoning length, and token efficiency. Taken together, these results show that CARE improves task performance while reallocating the reasoning budget more effectively during reinforcement learning.

\paragraph{Accuracy Reward}
As shown in the left panel of Fig.~\ref{fig:db_3t1}, the two models begin with similar reward values, but their trajectories gradually diverge as training proceeds. Video-R1 saturates after the mid stage and remains close to 0.67, whereas CARE continues to improve and stabilizes at a higher level near 0.70. This gap indicates that CARE maintains more effective optimization pressure after the baseline has largely plateaued.

\paragraph{Reasoning Length}
The middle panel shows a clear difference in how the two methods allocate response length during training. Video-R1 steadily grows to a roughly fixed reasoning budget and then fluctuates around that level. CARE, in contrast, first increases its reasoning length beyond the baseline and later reduces it markedly after the later training stage. This pattern suggests that CARE permits broader exploration when the policy is still immature, and then compresses the reasoning process once more reliable behavior has been acquired.

\paragraph{Token Efficiency}
The right panel further shows that the gain in reward is not achieved by simply maintaining longer outputs. We compute token efficiency as follows:
\begin{equation}
    \text{Token Efficiency} = \frac{\text{Accuracy}}{\text{Generated Tokens}} \times 1000,
\end{equation}
and observe that CARE rises substantially throughout training, eventually reaching a level far above Video-R1. Combined with the length reduction in the later stage, this result indicates that CARE improves both effectiveness and efficiency, producing more reward per generated token rather than relying on a larger token budget.

\begin{figure*}[t]
\centering
\includegraphics[width=\textwidth]{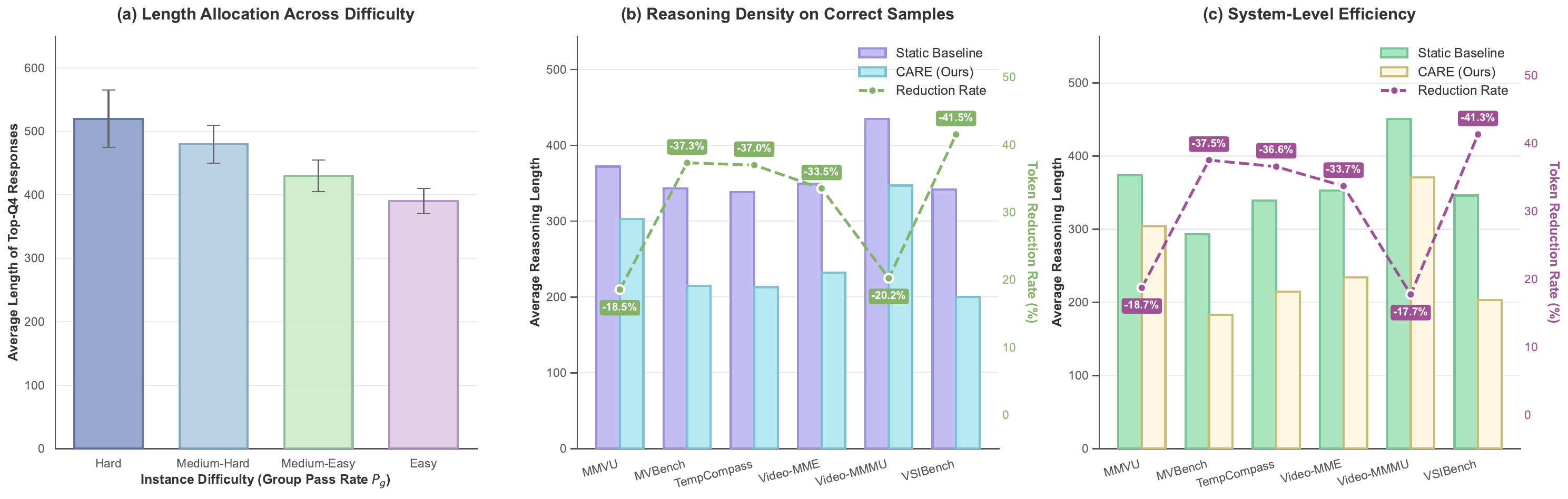}
\caption{Length allocation and efficiency analysis across difficulty groups and benchmarks. Left: average length of the longest-response group across difficulty levels. Middle: average reasoning length on correctly answered samples, with token reduction rates relative to the static baseline. Right: average reasoning length over all samples, with token reduction rates relative to the static baseline.}
\label{fig:bar_3t1}
\end{figure*}

\subsection{Length Allocation and Efficiency Analysis}
As summarized in Fig.~\ref{fig:bar_3t1}, the efficiency-related results are presented from three complementary views: length allocation across difficulty groups, reasoning length on correct samples, and reasoning length over the full evaluation set. Together, these results show that CARE reduces computation without collapsing the response budget uniformly across all cases.

\paragraph{Length Allocation Across Difficulty}
The left panel examines the longest-response group across different difficulty levels. A clear ordering remains: harder instances still receive longer responses, while easier instances are assigned shorter outputs. This result indicates that CARE does not simply truncate responses indiscriminately. Instead, it preserves a difficulty-sensitive allocation pattern and mainly compresses the portion of long responses that is less necessary for easier cases.

\paragraph{Reasoning Density on Correct Samples}
The middle panel restricts the analysis to correctly answered samples. Across all benchmarks, CARE consistently produces shorter correct responses than the static baseline, and the associated reduction rates remain substantial. Because answer correctness is already preserved in this comparison, the shorter responses indicate a denser reasoning process rather than an accuracy-efficiency trade-off.

\paragraph{System-Level Efficiency}
The right panel extends the comparison to the full evaluation set, including both correct and incorrect predictions. CARE again yields shorter average responses across benchmarks, with reduction rates that are broadly consistent with those observed on correct samples. This consistency suggests that the efficiency gain is not confined to a narrow subset of examples, but reflects a systematic shift toward more compact decoding behavior.

\subsection{Hyperparameter Sensitivity}

\begin{figure*}[!t]
\centering
\includegraphics[width=\textwidth]{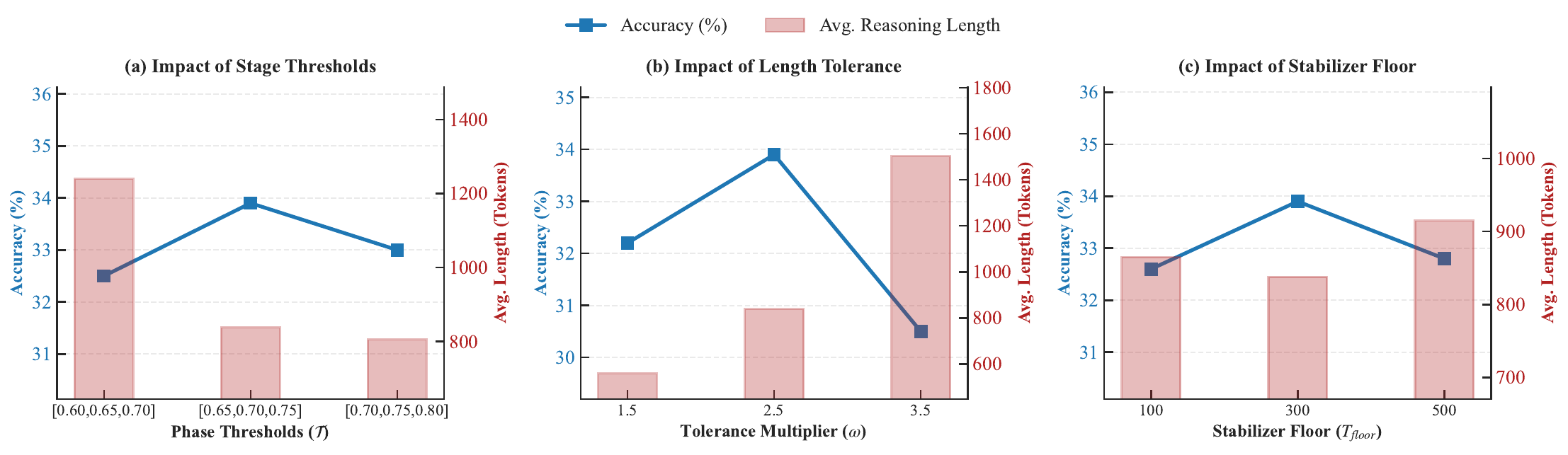}
\caption{Hyperparameter sensitivity of CARE with respect to stage thresholds, length tolerance, and stabilizer floor. The default setting achieves the strongest accuracy-length trade-off across all three factors.}
\label{fig:hyperparameter_sensitivity}
\end{figure*}

To examine whether CARE depends on a narrow parameter choice, we vary three representative controller hyperparameters that directly affect stage routing, response-length tolerance, and the minimum reasoning budget, while keeping all other settings fixed. As reported in Fig.~\ref{fig:hyperparameter_sensitivity}, both accuracy and average reasoning length are evaluated across variations. The default configuration in Table~\ref{tab:settings}, namely $\mathcal{T}=\{0.65,0.70,0.75\}$, $\omega=2.5$, and $T_{\mathrm{floor}}=300$, consistently provides the best or near-best accuracy while avoiding unnecessary token growth. This indicates that the adopted setting corresponds to a stable operating regime rather than a fragile point solution.

\paragraph{Effect of Stage Thresholds}
Adjusting the stage thresholds in either direction leads to weaker behavior. A more permissive setting, $\{0.60,0.65,0.70\}$, reduces accuracy and substantially increases the average reasoning length, indicating that overly early stage transitions make the routing signal less reliable and weaken the intended exploration-to-compression schedule. A more conservative setting, $\{0.70,0.75,0.80\}$, slightly shortens responses but also lowers accuracy, suggesting that delayed transitions underuse adaptive modulation. The default thresholds $\{0.65,0.70,0.75\}$ therefore provide the best balance between stable competence estimation and effective compression.

\paragraph{Effect of Length Tolerance}
The tolerance multiplier $\omega$ controls how strictly CARE reacts to response length deviations. When $\omega=1.5$, the controller becomes too restrictive and suppresses useful reasoning budget, leading to lower accuracy. When $\omega=3.5$, the constraint becomes too loose, causing a sharp increase in response length together with a clear accuracy drop. The middle setting $\omega=2.5$ achieves the highest accuracy with moderate reasoning length, showing that CARE benefits from a balanced tolerance that neither over-penalizes exploration nor permits redundant generation.

\paragraph{Effect of Stabilizer Floor}
The minimum reasoning floor $T_{\mathrm{floor}}$ plays a stabilizing role in early and uncertain cases. Setting $T_{\mathrm{floor}}=100$ yields lower accuracy, implying that the protected reasoning budget is insufficient for multi-step video reasoning. Increasing it to $500$ raises the average length without improving performance, which means the floor becomes overly conservative and preserves unnecessary computation. The default value $T_{\mathrm{floor}}=300$ offers the strongest trade-off, maintaining enough reasoning capacity while still allowing later compression.

\begin{figure}[!t]
\centering
\includegraphics[width=\columnwidth]{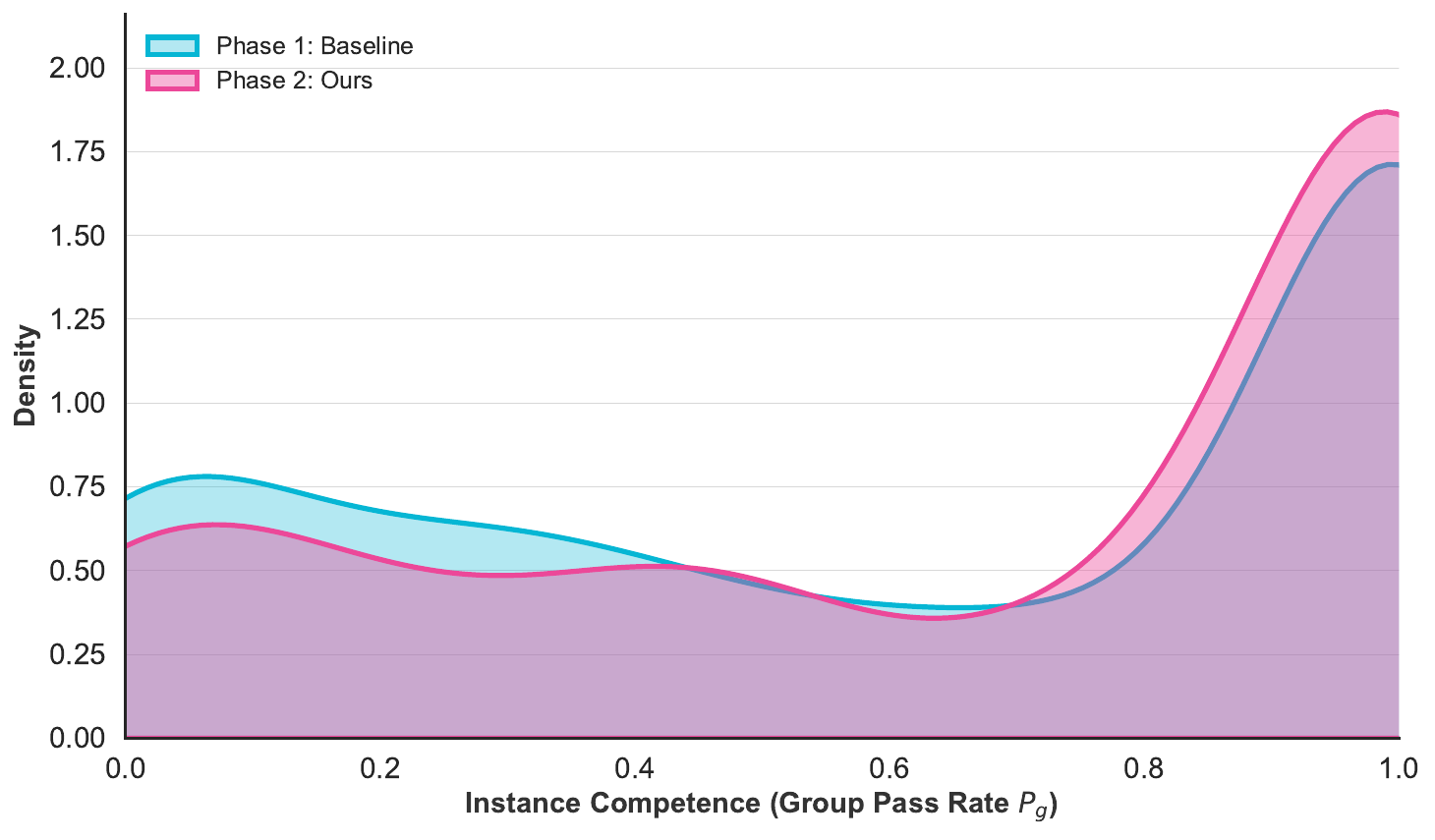}
\caption{Kernel density estimation of instance competence scores before and after CARE training. CARE shifts the distribution rightward and reduces the failure tail.}
\label{fig:competence_retention}
\end{figure}

\begin{figure}[!t]
\centering
\includegraphics[width=\columnwidth]{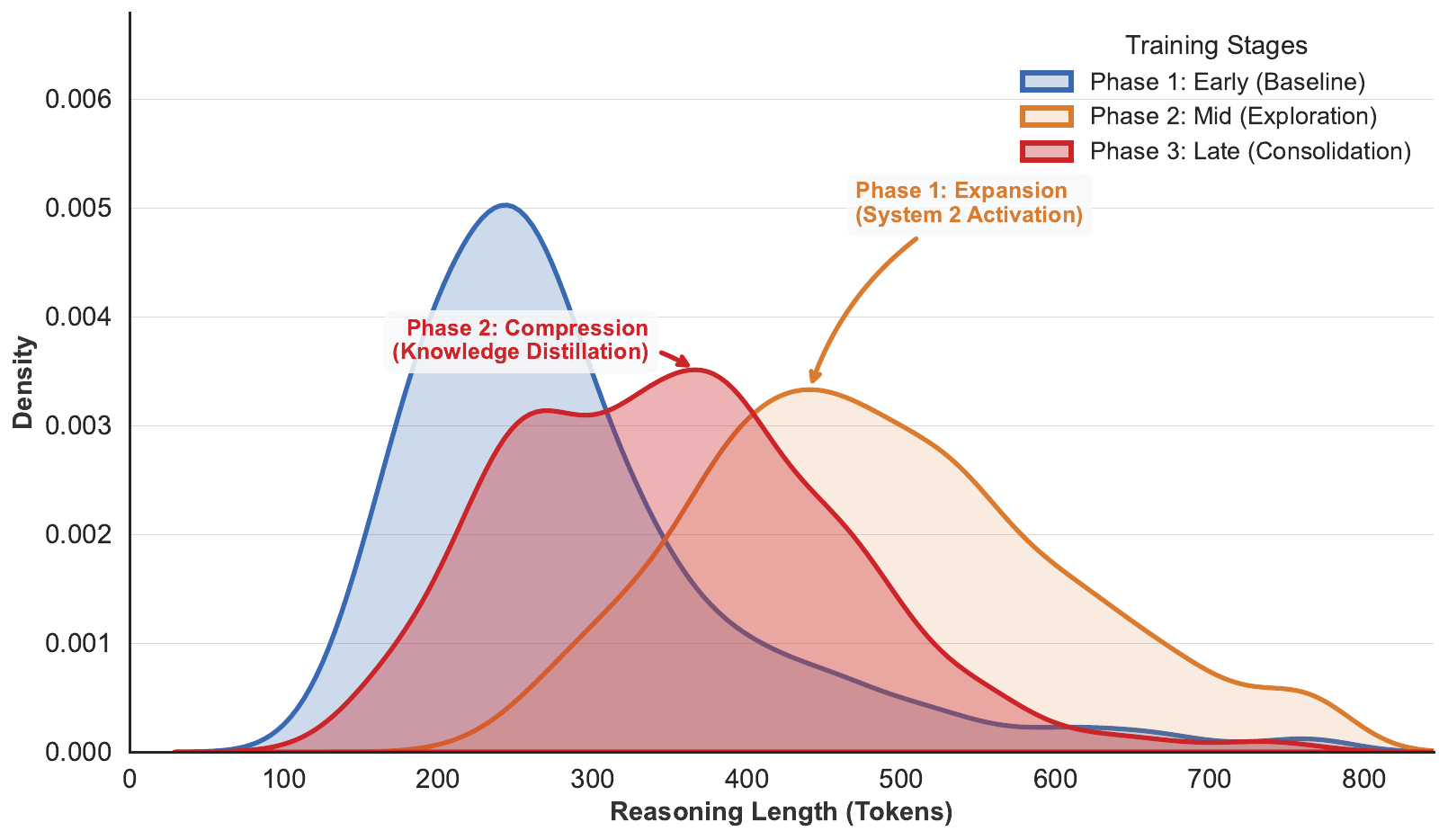}
\caption{Evolution of reasoning-length distributions across training phases. CARE first expands reasoning length and later compresses redundant responses.}
\label{fig:expansion_compression_cycle}
\end{figure}

\subsection{Expansion--Compression Cycle of Reasoning Length}
As visualized in Fig.~\ref{fig:expansion_compression_cycle}, this phenomenon can be explained by the expansion--compression cycle of reasoning lengths. During the early exploration phase, CARE shifts the length distribution rightward to enlarge the candidate reasoning space. In the late consolidation phase, the distribution shifts leftward, compressing redundant trajectories.

\begin{figure*}[!t]
\centering
\includegraphics[width=\textwidth]{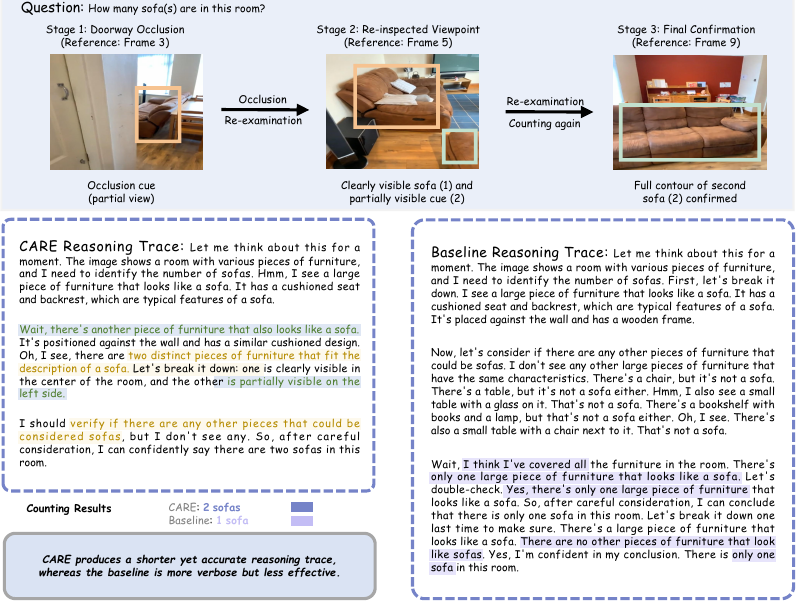}
\caption{Qualitative case study under viewpoint occlusion. CARE retains the partial cue from the early view, revisits the scene after re-examination, and correctly identifies two sofas once the second instance becomes fully visible. The baseline instead settles on a premature single-sofa interpretation and remains incorrect despite using a longer reasoning trace.}
\label{fig:case_study}
\end{figure*}

\subsection{Competence Retention Under Compression}
Crucially, as shown in Fig.~\ref{fig:competence_retention}, this compression does not degrade competence. Instead, the density of instances with near-perfect pass rates ($P_g \approx 1$) increases significantly under CARE. This proves that the observed length reduction reflects competence-aware compression, distilling internalized reasoning patterns into shorter, highly accurate responses without reverting to premature truncation.

\subsection{Case Study}
As presented in Fig.~\ref{fig:case_study}, a qualitative example illustrates a scenario where the target scene is initially observed under partial occlusion. CARE preserves the cue from the first view, re-examines the scene from a later viewpoint, and then updates the final count after the second sofa becomes fully visible. In contrast, the baseline commits early to a single-sofa interpretation and does not revise its conclusion after subsequent evidence appears. This example illustrates that CARE produces a shorter yet more effective reasoning process, remaining responsive to newly revealed visual evidence in multi-view video counting.

\section{Limitations}

Although CARE delivers consistent gains across a broad set of video reasoning evaluations, the current study should not be interpreted as claiming that one particular parameterization is tuned to achieve the best absolute result on every benchmark or metric. Our primary objective is to validate an automatic training principle: the reward preference over reasoning effort can evolve with the policy's estimated competence, so that exploration is protected when the model is still weak and redundant computation is compressed once the policy becomes stronger. Under a fixed backbone, a unified GRPO setup, and a shared hyperparameter schedule, the experiments support this auto-adjusting perspective through accuracy, efficiency, and trajectory analyses. Nevertheless, the absolute optimum may still vary with model scale, data mixture, or controller coefficients, and stronger benchmark-specific tuning may further improve the reported numbers. Therefore, the main takeaway of this work is the effectiveness of competence-aware automatic reward shaping as an interpretable training mechanism, rather than the universal optimality of the current hyperparameter setting.

\section{Conclusion}
In this paper, we presented CARE, a competence-aware reward shaping framework for adaptive reasoning length optimization in multimodal reasoning. Unlike conventional static length control strategies, CARE explicitly models reasoning efficiency as an evolving property during reinforcement learning, and dynamically adjusts reward incentives according to the model's competence. By combining a smoothed competence monitor, a stage-aware reward router, a difficulty-normalized effort scorer, and a posterior amplification mechanism, CARE enables the training process to transition from exploration-oriented long-form reasoning to efficiency-oriented concise reasoning. Extensive experiments on multiple video reasoning and general video understanding benchmarks demonstrated that CARE consistently improves reasoning accuracy, stabilizes reinforcement learning training, and enhances token efficiency. Further analyses showed that CARE induces a characteristic expansion-to-compression trajectory of reasoning length, allocates larger reasoning budgets to more difficult instances, and compresses redundant reasoning without sacrificing competence. These results suggest that adaptive reward shaping is a promising direction for training multimodal reasoning systems that are not only more accurate, but also more computationally efficient. In future work, we plan to extend competence-aware reward shaping to broader multimodal settings, including image, document, and embodied reasoning, and to explore finer-grained adaptive control signals beyond reasoning length alone.


\bibliographystyle{IEEEtran}
\bibliography{references}

\end{document}